# A Novel Transfer Learning Approach upon Hindi, Arabic, and Bangla Numerals using Convolutional Neural Networks

Abdul Kawsar Tushar, Akm Ashiquzzaman, Afia Afrin, and Md. Rashedul Islam[1]

Department of CSE, University of Asia Pacific, Dhaka, Bangladesh

`{tushar.kawsar, zamanashiq3, meghlaprottoy, rashed.cse}@gmail.com`

**Abstract.** Increased accuracy in predictive models for handwritten character recognition will open up new frontiers for optical character recognition. Major drawbacks of predictive machine learning models are headed by the elongated training time taken by some models, and the requirement that training and test data be in the same feature space and consist of the same distribution. In this study, these obstacles are minimized by presenting a model for transferring knowledge from one task to another. This model is presented for the recognition of handwritten numerals in Indic languages. The model utilizes convolutional neural networks with backpropagation for error reduction and dropout for data overfitting. The output performance of the proposed neural network is shown to have closely matched other state-of-the-art methods using only a fraction of time used by the state-of-the-arts.

## 1 Introduction

Knowing how to cook fish helps while cooking chicken. Solving some mathematical problems enhances the ability to solve other similar problems. Once the basic circuitry of a small cell phone is known, it gets easier to explain the mechanism of similar other electronic devices. Knowledge is a dynamic horizon of the repertoire of human beings. Learning a new technique and applying it to solve different related problems - this is the natural way of learning. For example, when we learn how to differentiate between rotten potato and fresh potato we intuitively learn the basics of identifying a rotten vegetable, be it a tomato or potato!

Such observations lead us to a specialized learning method - where previous experiences and stored knowledge are used to resolve new tasks and problems. This tech-nique is known as *transfer learning*. It is an important subsection of the field of machine learning (ML). Transfer is one of the most important elements of human learning mechanism which indicates that one major application of transfer learning is to deploy it in neural networks, as neural network is modeled after the human brain and nervous system. Several experiments and research works agree with this. For example, experimental results obtained from [1] provide evidence that transferring knowledge across related tasks helps the learner experience more and generalize better. An inductive transfer mechanism namely multitask learning has been demonstrated in [2] which improves generalization by learning tasks in parallel while using a shared representation. A similar approach, self-taught learning, has been adopted in [3] which works with unlabeled data sets.

---
[1] Corresponding Author

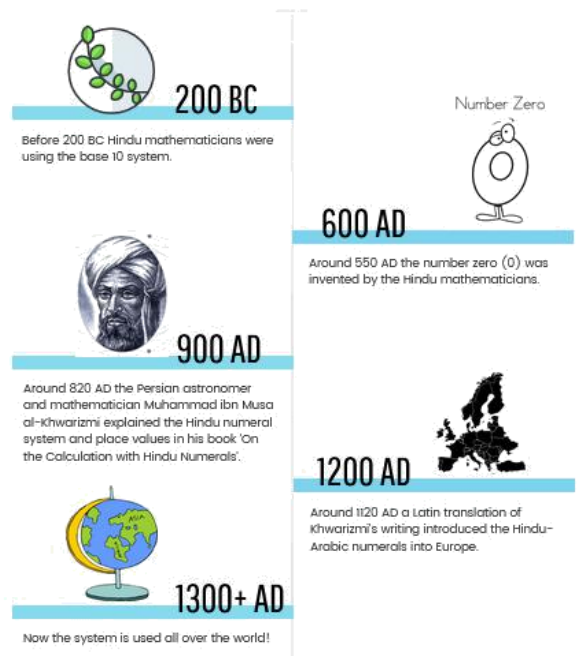

**Fig. 1.** Brief History of the Modern Number System

Modern decimal system is a descendant of the Hindu-Arabic numeral system, invented by the Indian mathematicians between the 1st and 4th centuries. Very early use of the place value system is seen in the *Bakhshali manuscript*, the oldest extant manuscript in Indian mathematics [4]. Though the actual date of the composition of the manuscript is a matter of debate, the language used in the manuscript indicates that it could not have been composed any later than 400 AD. The oldest known mention of numeral zero ("0") and the decimal positional system has been found in the *Lokavibhaga*, a *Jain* cosmological text translated from a *Prakrit* original internally dated to AD 458 [5, 6]. By the 9th century the system was adopted in Arabic mathematics and later introduced to Europe by the High Middle Ages (around 12th century) [7]. The book *On the Calculation with Hindu Numerals*, written in about 825 AD in Arabic by a Persian astronomer and mathematician, Muhammad Ibn Musa al-Khwarizmi introduced the Indian numeral system to the Arabian Peninsula. Later in the 12th century, the Western world got introduced to this system through the Latin translations of his work [8]. Thus, the Indian numerals form the basis of European number systems which are now broadly used worldwide. However, this long journey of numbers from India to Europe through Arab was not as simple as it sounds. The Eastern and Western parts of the Arabic world adopted the basic Indian numeral system differently and the European digits that we observe nowadays are descendant of the Western Arabic glyphs. On the other hand, the Eastern Arabic numerals (also known as the "Indic numerals") spread among many countries to the East of the Arab world. Fig. 1 shows the entire historical journey of the modern number system using a simple timeline. Nonetheless, the history of invention and evaluation of modern number system clearly stipulates that Hindu-Arabic numeral system is the precursor of modern numerals. This observation indicates a high correlation among different numeral systems which is the primary motivation behind our work.

This paper proposes a deep learning model appropriate for training and transferring the knowledge to a later task. Later transfer learning techniques are utilized to compare reduced training time and recognition accuracy where prediction accuracy for image numerals are scored. The model is trained individually with three different existing numeral systems in image forms, namely - Bangla, Urdu, and Hindi; all of these originated from the Indic numerals under Indic languages which are used by a considerable percentage of world's population. All these three numeral systems, having a fairly recent common predecessor, are highly correlated with each other which consequently makes it possible to apply transfer learning, as stated above.

The rest of this paper is structured as follows: Section 2 introduces the concept of deep neural networks (DNN) with which the model is formulated. Section 3 outlines the proposed model which follows the explanation for transfer learning. Section 4 provides details about the experimentation and relevant dataset, and discusses about the results. Then Section 5 concludes the paper.

## 2    Deep Neural Network

This section presents an overview of the deep learning system that will be used in the proposed model. Deep neural network is one type of artificial neural network (ANN) - a computing system inspired by the biological neural networks. Whereas a simple predictive algorithm tries to mimic the mapping between input and output variables, ANN has a unique characteristic of creating transient states through the *artificial neurons* which are the basic building blocks of ANN. What constitutes the main difference between human brain and simple machine is the creativity and decision taking capability. ANN is the first step of modern technology to eradicate this "little" difference.

Artificial neurons organized in several layers form an artificial neural network. A DNN will presumably have more layers than a simple ANN. Though the number of layers is not fixed, it is usually no less than four for DNN. Among various types of DNN, the Convolutional neural networks (CNN) are widely used for processing visual and other two-dimensional data [9, 10].

## 3    Proposed Method

This section outlines a brief technical discussion on transfer learning and then describes the proposed technique that is central to this study.

### 3.1    Transfer Learning

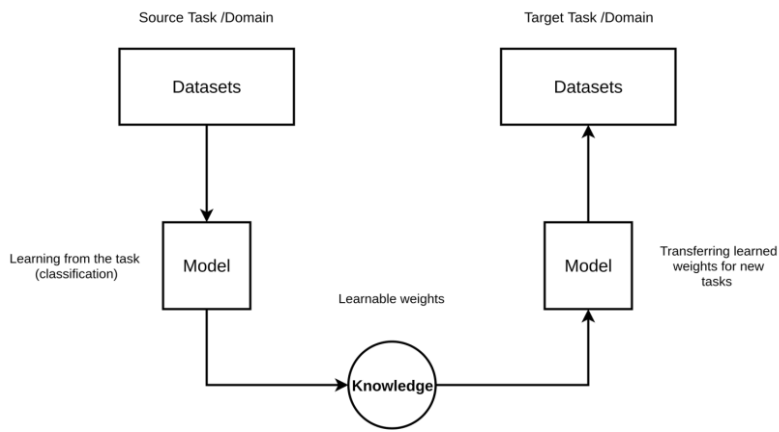

**Fig. 2.** Transfer Learning Flow Diagram.

ML and data mining technologies have introduced a revolutionary change in the field of knowledge engineering including classification, clustering and regression [11, 12]. However, a major drawback of these renowned techniques is most of the ML algorithms perform well only when training and test data are drawn from the same feature space and follow the same distribution. When this *identical distribution* assumption does not hold, most of the ML models need to be rebuilt from scratch with new sets of training data which is undoubtedly a time-consuming approach. Furthermore, large amounts of hand-crafted, structured training data are required for this purpose, which is expensive. Thus, despite being a successful learning technique, prevalent ML algorithms may fail in some real-world situations [13]. In such cases, *knowledge transfer* can improve the learning performance, as transferring knowledge between different but similar problem domains eliminates the extra burden of recollecting training data and rebuilding the model from scratch. One such example is image clustering where heterogeneous data from different feature spaces are used for training purpose. A novel method for image clustering using heterogeneous transfer learning technique has been proposed in [14]. Similar techniques have been devised in [15] and [16] for the task of classifying Chinese text documents using another collection of English texts as training data.

Another major application of transfer learning is when there is a chance of training data to be outdated with time. One such example is *Web mining* [17], where the Web data used in training a particular Web-page classification model can be easily outdated afterwards, as the topics on the web change frequently. In general, training is acquired from a source task and infused into the model. The knowledge from the model is then carried over to another model working on a new task in same or different domain. The target task is then re-trained while utilizing the transferred knowledge to its advantage in reduction of time and increase in accuracy. Fig. 2 depicts these steps between the source and target tasks.

### 3.2 Proposed Model Architecture

The whole model can be divided into two main clusters depending on their tasks: [9] and [18] use CNN layers as unsupervised feature vector extractors and fully connected layers as classifiers in their studies. The current study follows the same footprint. The feature extractor cluster consists of four separate convolutional layers with variable output as well as kernel size 3x3. Each convolution layer learns increasingly complex patterns or features of the input image data and gradually shifts closer toward recognition of the same. After the last convolution layer, a max pooling layer with 2x2 sized kernel and 2 sized strides in both axes is inserted in order to increase weight of the most important features. The classification cluster has three fully connected neural network layers with same variable output numbers. The fully connected layers are used as classifiers which act upon complex features learned in the convolution layers. All of the layers have exponential linear unit (ELU) as activation functions, except for the last layer; the endmost layer has a Softmax function for output [19]. ELU is used to overcome vanishing gradient problem during training [20]. The convolution and fully connected layers are insured with dropout function to reduce overfitting [21]. Details of the neural network layers are described in the Table 1.

The whole model is trained from scratch with numerals from the mentioned three scripts, separately. The first phase of the process sees the models being trained for 300 epochs and the best resulting weights subsequently saved in each case. In the second phase, all of the feature extraction cluster, i.e. the CNN layers are halted for weight update due to overfitting concerns. At the same time, layers in the classification cluster get to learn to detect different digits using feature vectors obtained from recognition process of numerals of another language. All simulations are performed on a computer hosting Intel i3 7th generation processor with 8GB RAM. Nvidia GTX-1050Ti 4GB DDR-5 GPU is used for CUDA accelerated computing.

**Table 1.** Architecture of Deep Neural Network

| Layer | Number of Neurons | Activation Function Used | Kernel Size |
|---|---|---|---|
| Convolutional | 64 | ELU | 3x3 |
| Convolutional | 64 | ELU | 3x3 |
| Convolutional | 64 | ELU | 3x3 |
| Convolutional | 32 | ELU | 3x3 |
| Max Pooling | N/A | N/A | 2x2 |
| Fully Connected | 512 | ELU | N/A |
| Fully Connected | 256 | ELU | N/A |
| Fully Connected | 128 | ELU | N/A |
| Output | 10 | Softmax | N/A |

## 4    Experiments

This section presents a brief discussion on the data sets that are used in the experiments and then examines the results of these experiments.

### 4.1    Dataset

In this study, all of the images chosen are samples of handwritten numerals collected form a range of writers. Bangla handwritten numerical dataset CMATERDB 3.1.1, Hindi (Devanagari) handwritten numerical dataset CMATERDB 3.2.1, and Urdu (Arabic) handwritten numerical dataset CMATERDB 3.3.1 are chosen for this study [22]. All three of these datasets contain handwritten 32x32 pixels that are binary-scanned and segmented samples of the digits. Table 2 shows brief details of the datasets.

The size and quality of training dataset greatly influence the accuracy and correctness of any ML algorithm. For deep learning, this effect is even more prominent, as convergence rate of any neural network is highly dependent on the model complexity and availability of data. In this study, we have used three datasets containing handwritten digits of different languages sharing a common ancestor. The images are inverted before feeding into CNN. As a result, the numerals are in white foreground on the backdrop of black background. Edges are a very important feature in character recognition, converting the background black along with white foreground makes the edge detection slightly easier [23].

Table 2. Brief Details of the Dataset.

| ID | Lingual Origin of The Numerals | Training Samples | Digits/Classes |
|---|---|---|---|
| 1 | Urdu | 3000 | 10 |
| 2 | Bangla | 6000 | 10 |
| 3 | Hindi | 3000 | 10 |

### 4.2 Results and Discussion

An objective of this study is to verify the hypothesis that numerals of the same origins will be easier of detect as they have shared the morphological common traits. It will be difficult to estimate competitiveness of performance for transfer learning solely by observing similar features for both source and targeted domains. In contrast, key insights will be gained by considering the time of training deep learning models. For this purpose, firstly, Table 3 shows the comparison of the three tasks, i.e. separate simulations on three datasets described in Section 3. The table demonstrates the learning times, or the numbers of iterations over the whole dataset the model needs to classify with state-of-the-art correctness. For each of the datasets used, the accuracy is observed to have been achieved at around 200th epochs on average. Note that after the best accuracy is achieved, the model for that epoch is saved and used in subsequent epochs for the same dataset. For another one of the datasets, a fresh model is used where weights and biases are again initialized. Therefore, parameters of these three models are in no way dependent on each other.

Table 3. Standalone Training Performance on Datasets.

| Dataset Used for Training | Best Accuracy Achieved | Number of Epochs | Best Accuracy in Epoch Number |
|---|---|---|---|
| Urdu | 99.30% | 300 | 230 |
| Bangla | 99.40% | 300 | 216 |
| Hindi | 99.26% | 300 | 189 |

After that, weights and biases from the three isolated runs of the model with each of the datasets are saved and re-used for a different task, different from the source task. At this stage, the feature extraction cluster of the models, as mentioned in Section 4, are frozen for weight update, otherwise the main objective of transfer learning would not be ensured. 100 epochs are allocated to each of the three runs at this stage. Table 4 shows the task combinations with which transfer learning experiments are executed. We highlight three things in this result: the best accuracy achieved with the transferred model, epoch number where this best accuracy is achieved, and the accuracy of that run as recorded after ten initial epochs are completed. It is observed form Table 4 that, even though the feature extraction layers are trained on a different task in an unsupervised manner, the classification accuracy is on par with state-of-the-art methods mentioned in Table 3. It is also evident that all the transferred models are capable of instantaneous recognition of digits from a related but different numeral script. This power is lent to the transferred model from the intuitive explanation of transfer learning mentioned in Section 1. Another remarkable insight gained from this experiment is that, in all the runs, over 92% accuracy is achieved after only 10 epochs. It is a very short time compared to the original model mentioned in Table 3 where it took 300 epochs to train them sufficiently enough. A point of note here is that the accuracy could have improved further, had the models been run for more than the 100 epochs allocated. Furthermore, when a task does not have sufficient labeled training data, the proposed approach will be of significant help if the model is pre-trained on a related but separate dataset.

**Table 4.** Transfer Learning Comparisons.

| Source Task | Destination Task | Best Accuracy Achieved | Best Accuracy in Epoch Number | Accuracy After 10 Epochs |
|---|---|---|---|---|
| Urdu | Bangla | 96.99% | 42 | 93.90% |
| Bangla | Urdu | 97.79% | 48 | 97.12% |
| Hindi | Bangla | 98.66% | 38 | 95.45% |
| Urdu | Hindi | 95.88% | 77 | 92.11% |
| Bangla | Hindi | 98.57% | 52 | 93.67% |
| Hindi | Urdu | 98.57% | 64 | 92.44% |

Results of the transfer learning process can, in part, be explained through the morphological similarities of the digits among the languages under discussion. However, the feature extraction process of the CNN is in itself an unsupervised operation. Moreover, where the model will be applied after transfer is not known from before to the earlier task model. Therefore, it cannot be said that before transferring, the model learns the important features it recognizes in the previous task to better adapt to the task after transfer. Interestingly, this phenomenon partially explains the discrepancy in accuracy in transferring skills from one language to another and vice versa in Table 4. The competitive results obtained from the proposed transfer learning process, taken along with the limited amount of re-training time, can be termed as a very competitive process.

## 5    Conclusion

In this research, a model for transferring knowledge from one character recognition task to another in proposed. The model is used for recognition of handwritten numerals in Indic scripts used by a considerable percentage of world's population. A novel form of convolutional neural network for transfer learning with curtailed time and competitive accuracy is discussed for this purpose. Independence of model-specific training has helped to reduce the re-training time in the target task significantly. Furthermore, by diminishing the effect of overfitting in the proposed model, very competitive accuracy is achieved via experimentation. As a result, performance for predictive models for numeral recognition can now have prediction scores or performance gains on par with state-of-art methods in shorter time, which can lead to future breakthroughs in numeral recognition tasks in specific and transfer learning in general.

**Acknowledgement**

The authors would like to thank the department of Computer Science and Engineering, University of Asia Pacific for supporting this research in various ways. Abdul Kawsar Tushar and Akm Ashiquzzaman contributed equally to this work.